\documentclass[10pt,twocolumn,letterpaper]{article}

\usepackage{cvpr}
\usepackage{times}
\usepackage{epsfig}
\usepackage{graphicx}
\usepackage{amsmath}
\usepackage{amssymb}
\usepackage{bm}
\usepackage{subcaption}
\usepackage{booktabs}
\usepackage{multirow}
\usepackage{tabularx}
\usepackage{makecell}

\DeclareMathOperator*{\argmin}{arg\,min}

\usepackage[breaklinks=true,bookmarks=false,colorlinks]{hyperref}

\cvprfinalcopy 

\def\cvprPaperID{4445} 

\begin{document}

\title{AANet: Adaptive Aggregation Network for Efficient Stereo Matching}

\author{Haofei Xu \quad Juyong Zhang\thanks{Corresponding author} \\
University of Science and Technology of China \\
{\tt\small xhf@mail.ustc.edu.cn, juyong@ustc.edu.cn}
}

\maketitle

\begin{abstract}

Despite the remarkable progress made by learning based stereo matching algorithms, one key challenge remains unsolved. Current state-of-the-art stereo models are mostly based on costly 3D convolutions, the cubic computational complexity and high memory consumption make it quite expensive to deploy in real-world applications.
In this paper, we aim at completely replacing the commonly used 3D convolutions to achieve fast inference speed while maintaining comparable accuracy. To this end, we first propose a sparse points based intra-scale cost aggregation method to alleviate the well-known edge-fattening issue at disparity discontinuities. Further, we approximate traditional cross-scale cost aggregation algorithm with neural network layers to handle large textureless regions. Both modules are simple, lightweight, and complementary, leading to an effective and efficient architecture for cost aggregation. With these two modules, we can not only significantly speed up existing top-performing models (e.g., $41\times$ than GC-Net, $4\times$ than PSMNet and $38\times$ than GA-Net), but also improve the performance of fast stereo models (e.g., StereoNet). We also achieve competitive results on Scene Flow and KITTI datasets while running at 62ms, demonstrating the versatility and high efficiency of the proposed method. Our full framework is available at \url{https://github.com/haofeixu/aanet}.

\end{abstract}

\section{Introduction}

Estimating depth from stereo pairs is one of the most fundamental problems in computer vision \cite{scharstein2002taxonomy}. The key task is to find spatial pixel correspondences, i.e., stereo matching, then depth can be recovered by triangulation. Efficient and accurate stereo matching algorithms are crucial for many real-world applications that require fast and reliable responses, such as robot navigation, augmented reality and autonomous driving.

Traditional stereo matching algorithms generally perform a four-step pipeline: matching cost computation, cost aggregation, disparity computation and refinement, and they can be broadly classified into global and local methods \cite{scharstein2002taxonomy}. Global methods usually solve an optimization problem by minimizing a global objective function that contains data and smoothness terms \cite{sun2003stereo, kolmogorov2001computing}, while local methods only consider neighbor information \cite{yoon2006adaptive, hosni2012fast}, making themselves much faster than global methods \cite{min2011revisit, scharstein2002taxonomy}. Although significant progress has been made by traditional methods, they still suffer in challenging situations like textureless regions, repetitive patterns and thin structures.

\begin{figure}[!t]
    \centering
    \begin{subfigure}{0.32\linewidth}
        \includegraphics[width=\linewidth]{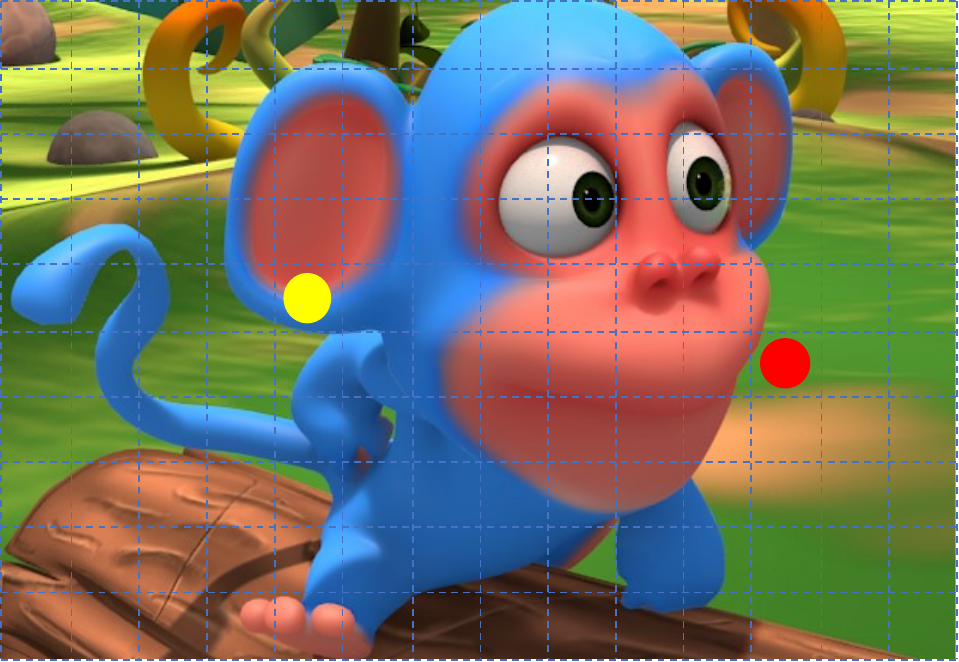}
        \caption{}
        \label{}
    \end{subfigure}
    \begin{subfigure}{0.32\linewidth}
        \includegraphics[width=\linewidth]{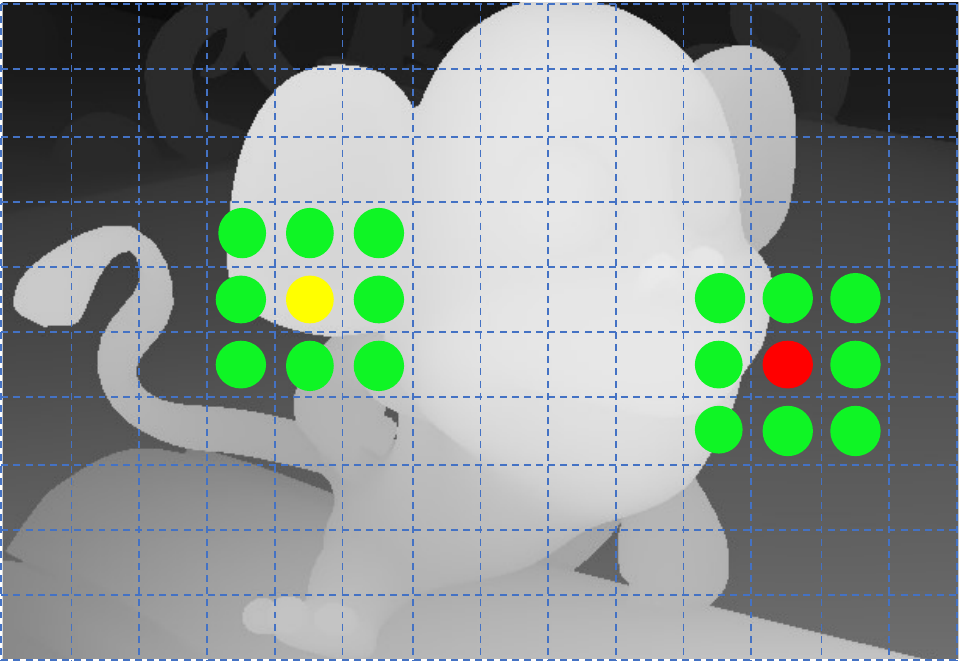}
        \caption{}
        \label{}
    \end{subfigure}
    \begin{subfigure}{0.32\linewidth}
        \includegraphics[width=\linewidth]{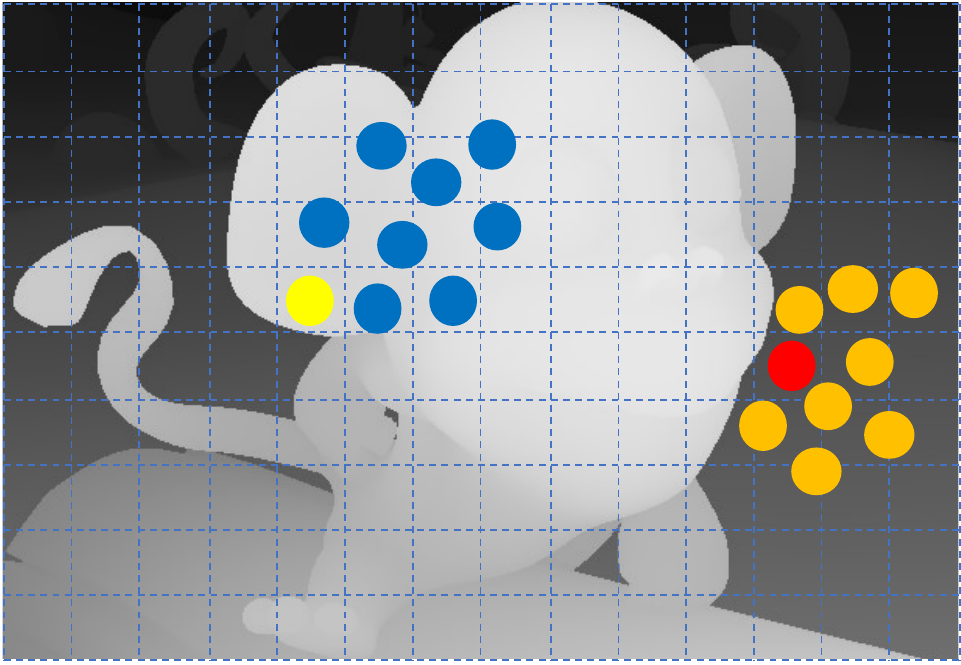}
        \caption{}
        \label{}
    \end{subfigure}
    \caption{Illustration of the sampling locations in regular convolution based cost aggregation methods and our proposed approach, where the yellow and red points represent the locations for aggregation. (a) left image of a stereo pair. (b) fixed sampling locations in regular convolutions, also the aggregation weights are spatially shared.
    (c) adaptive sampling locations and position-specific aggregation weights in our approach. The background in (b) and (c) is ground truth disparity.}
    \label{fig:adaptive_sampling}
\end{figure}

Learning based methods make use of deep neural networks to learn strong representations from data, achieving promising results even in those challenging situations. 
DispNetC \cite{mayer2016large} builds the first end-to-end trainable framework for disparity estimation, where a correlation layer is used to measure the similarity of left and right image features. GC-Net \cite{kendall2017end} takes a different approach by directly concatenating left and right features, and thus 3D convolutions are required to aggregate the resulting 4D cost volume. PSMNet \cite{chang2018pyramid} further improves GC-Net by introducing more 3D convolutions for cost aggregation and accordingly obtains better accuracy. Although state-of-the-art performance can be achieved with 3D convolutions, the high computational cost and memory consumption make it quite expensive to deploy in practice (for example, PSMNet costs about 4G memory and 410ms to predict a KITTI stereo pair even on high-end GPUs). The recent work, GA-Net \cite{zhang2019ga}, also notices the drawbacks of 3D convolutions and tries to replace them with two guided aggregation layers. However, their final model still uses fifteen 3D convolutions. 

To this end, a motivating question arises: \textit{How to achieve state-of-the-art results without any 3D convolutions while being significantly faster?} 
Answering this question is especially challenging due to the strong regularization provided by 3D convolutions. In this paper,
we show that by designing two effective and efficient modules for cost aggregation,
competitive performance can be obtained on both Scene Flow and KITTI datasets even \textit{with simple feature correlation \cite{mayer2016large} instead of concatenation \cite{kendall2017end}}.

Specifically, we first propose a new sparse points based representation for intra-scale cost aggregation.
As illustrated in Fig.~\ref{fig:adaptive_sampling}, a set of sparse points are adaptively sampled to locate themselves in regions with similar disparities, alleviating the well-known edge-fattening issue at disparity discontinuities \cite{scharstein2002taxonomy}. Moreover, such representation is flexible to sample from a large context while being much more efficient than sampling from a large window, an essential requirement for traditional local methods to obtain high-quality results \cite{min2011revisit}. We additionally learn content-adaptive weights to achieve position-specific weighting for cost aggregation, aiming to overcome the inherent drawback of spatial sharing nature in regular convolutions. We implement the above ideas with deformable convolution \cite{zhu2019deformable}.

We further approximate traditional cross-scale cost aggregation algorithm \cite{zhang2014cross} with neural network layers by constructing multi-scale cost volumes in parallel and allowing adaptive multi-scale interactions, producing accurate disparity predictions even in low-texture or textureless regions.

These two modules are simple, lightweight, and complementary, leading to an efficient architecture for cost aggregation.
We also make extensive use of the key ideas in the feature extraction stage, resulting in our highly efficient and accurate Adaptive Aggregation Network (AANet). For instance, we can outperform existing top-performing models on Scene Flow dataset, while being significantly faster, e.g., $41\times$ than GC-Net\cite{kendall2017end}, $4\times$ than PSMNet \cite{chang2018pyramid} and $38\times$ than GA-Net \cite{zhang2019ga}. Our method can also be a valuable way to improve the performance of fast stereo models, e.g., StereoNet \cite{khamis2018stereonet}, which are usually based on a very low-resolution cost volume to achieve fast speed, while at the cost of sacrificing accuracy. We also achieve competitive performance on KITTI dataset while running at 62ms, demonstrating the versatility and high efficiency of the proposed method.

\begin{figure*}[t]
    \centering
    \includegraphics[width=0.8\linewidth]{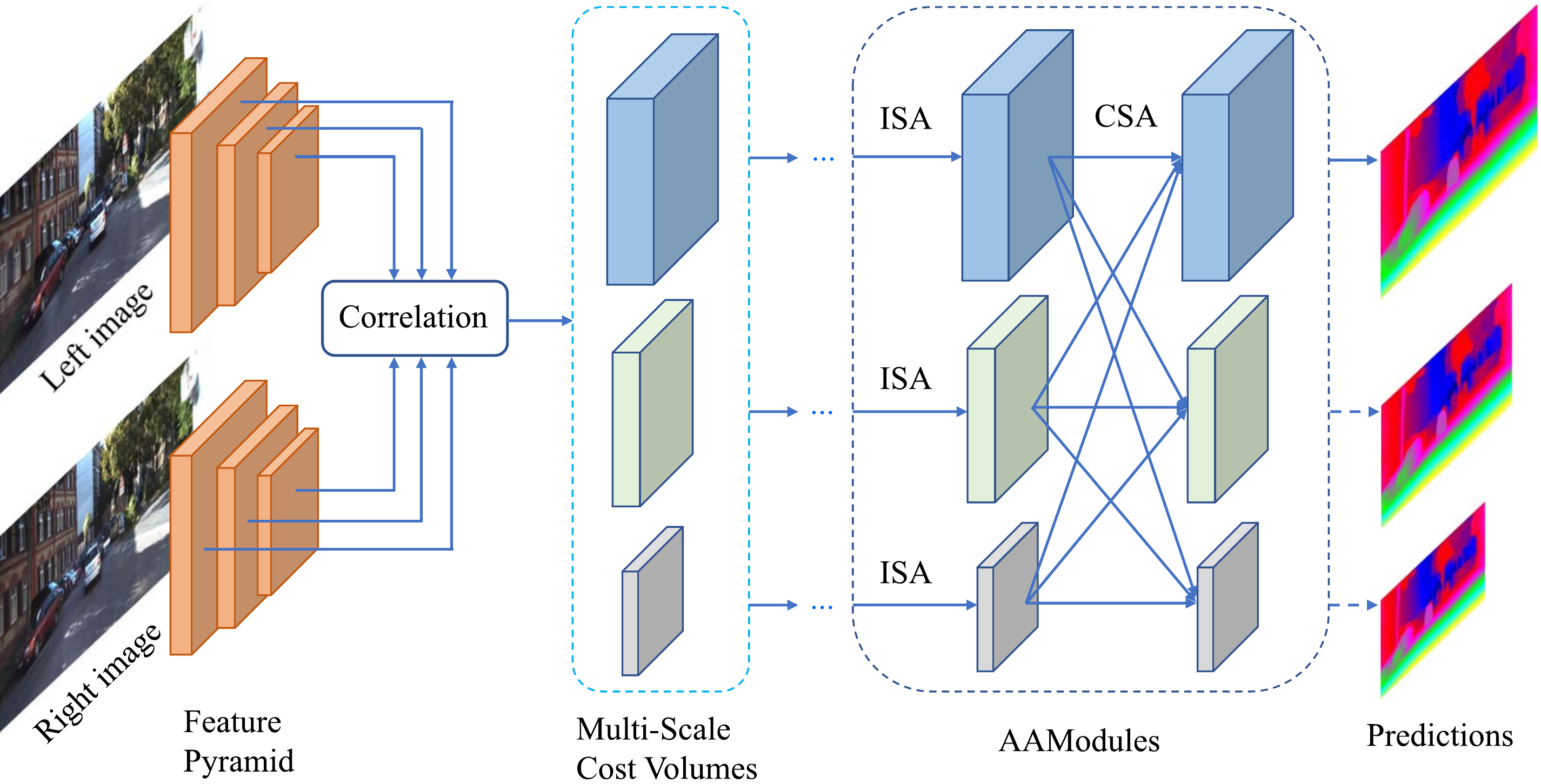}
    \caption{Overview of our proposed Adaptive Aggregation Network (AANet). Given a stereo pair, we first extract downsampled feature pyramid at $1/3, 1/6$ and $1/12$ resolutions with a shared feature extractor. Then multi-scale cost volumes are constructed by correlating left and right features at corresponding scales. The raw cost volumes are aggregated by six stacked Adaptive Aggregation Modules (AAModules), where an AAModule consists of three Intra-Scale Aggregation (ISA, Sec.~\ref{sec:intra}) modules and a Cross-Scale Aggregation (CSA, Sec.~\ref{sec:inter}) module for three pyramid levels. Next multi-scale disparity predictions are regressed. Note that the dashed arrows are only required for training
    and can be removed for inference. Finally the disparity prediction at $1/3$ resolution is hierarchically upsampled/refined to the original resolution. For clarity, the refinement modules are omitted in this figure, see Sec.~\ref{sec:aanet} for details.
    }
    \label{fig:overview}
\end{figure*}

\section{Related Work}

This section reviews the most relevant work to ours.

{\bf Local Cost Aggregation.}
Local stereo methods (either traditional \cite{yoon2006adaptive, hosni2012fast} or 2D/3D convolution based methods \cite{mayer2016large, kendall2017end}) usually perform window based cost aggregation:
\begin{equation}
    \label{eq:local}
    \Tilde{\bm C}(d, \bm p) = \sum_{\bm q \in N(\bm p)} w(\bm p, \bm q) \bm C (d, \bm q),
\end{equation}
where $\Tilde{\bm C} (d, \bm p)$ denotes the aggregated cost at pixel $\bm p$ for disparity candidate $d$, pixel $\bm q$ belongs to the neighbors $N(\bm p)$ of $\bm p$, $w(\bm p, \bm q)$ is the aggregation weight and $\bm C (d, \bm q)$ is the raw matching cost at $\bm q$ for disparity $d$. Despite the widespread and successful applications of local methods, they still have several important limitations. First and foremost, the fundamental assumption made by local methods is that all the pixels in the matching window have similar disparities. However, this assumption does not hold at disparity discontinuities, causing the well-known edge-fattening issue in object boundaries and thin structures \cite{scharstein2002taxonomy, min2011revisit}. As a consequence, the weighting function $w$ needs to be designed carefully to eliminate the influence of pixels that violate the smoothness assumption \cite{hosni2012fast, yoon2006adaptive}. While learning based methods automatically learn the aggregation weights from data, they still suffer from the inherent drawback of regular convolutions: weights are spatially shared, thus making themselves content-agnostic. 
Moreover, a large window size is often required to obtain high-quality results \cite{min2008cost, min2011revisit}, leading to high computational cost. 
Some works have been proposed to address the limitations of fixed rectangular window, e.g., using varying window size \cite{okutomi1992locally}, multiple windows \cite{hirschmuller2002real}, or unconstrained shapes \cite{boykov1998variable}. 

Different from existing methods, we propose a new sparse points based representation for cost aggregation. This representation is also different from \cite{min2011revisit}, in which sparse points inside the matching window are regularly sampled to reduce the computational complexity. In contrast, our proposed sampling mechanism is completely unconstrained and adaptive, providing more flexibility than the regular sampling in \cite{min2011revisit}. We also learn additional content-adaptive weights to enable position-specific weighting in contrast to the spatial sharing nature of regular convolutions.


{\bf Cross-Scale Cost Aggregation.}
Traditional cross-scale cost aggregation algorithm \cite{zhang2014cross} reformulates local cost aggregation from a unified optimization perspective, and shows that by enforcing multi-scale consistency on cost volumes, the final cost volume is obtained through the adaptive combination of the results of cost aggregation performed at different scales. 
Details are provided in the supplementary material. 
We approximate this conclusion with neural network layers in an end-to-end manner. Different from existing coarse-to-fine approaches \cite{tonioni2019real,yin2019hierarchical,sun2018pwc},
we build multi-scale cost volumes in parallel and allow adaptive multi-scale interactions. 
Our cross-scale aggregation architecture is also different from the very recent work \cite{wu2019semantic}, in which multi-scale cost volumes are also constructed. However, \cite{wu2019semantic} fuses the cost volumes from the lowest level to the higher ones hierarchically, while ours aggregates all scale cost volumes simultaneously based on the analysis in \cite{zhang2014cross}.

{\bf Stereo Matching Networks.}
Existing end-to-end stereo matching networks can be broadly classified into two categories: 2D and 3D convolution based methods. They mainly differ in the way that cost volume is constructed. 2D methods \cite{mayer2016large,liang2018learning,tonioni2019real} generally adopt a correlation layer \cite{mayer2016large} while 3D methods \cite{kendall2017end,chang2018pyramid,nie2019multi,zhang2019ga,chabra2019stereodrnet} mostly use direct feature concatenation \cite{kendall2017end}. An exception to concatenation based 3D methods is \cite{guo2019group}, in which group-wise correlation is proposed to reduce the information loss of full correlation \cite{mayer2016large}. In terms of performance, 3D methods usually outperform 2D methods by a large margin on popular benchmarks (e.g., Scene Flow \cite{mayer2016large} and KITTI \cite{menze2015object}), but the running speed is considerably slower. In this paper, we aim at significantly speeding up existing top-performing methods while maintaining comparable performance.
The very recent work, DeepPruner \cite{duggal2019deeppruner}, shares a similar goal with us to build efficient stereo models. They propose to reduce the disparity search range by a differentiable PatchMatch \cite{barnes2009patchmatch} module, and thus a compact cost volume is constructed. In contrast, we aim at reducing the sampling complexity and improving the sampling flexibility in cost aggregation, which works on different aspects, and both methods can be complementary to each other. 

{\bf Deformable Convolution.}
Deformable convolution \cite{dai2017deformable, zhu2019deformable} is initially designed to enhance standard convolution's capability of modeling geometric transformations, and commonly used as backbone for object detection and semantic/instance segmentation tasks. We instead take a new perspective of traditional stereo methods and propose an adaptive sampling scheme for efficient and flexible cost aggregation. Since the resulting formulation is similar to deformable convolution, we adopt it in our implementation. 


\section{Method}

Given a rectified image pair $\bm I_l$ and $\bm I_r$, we first extract downsampled feature pyramid $\{\bm F_l^s \}_{s=1}^S$ and $\{\bm F_r^s \}_{s=1}^S$ with a shared feature extractor, where $S$ denotes the number of scales, $s$ is the scale index, and $s=1$ represents the highest scale. Then multi-scale 3D cost volumes $\{\bm C^s \}_{s=1}^S$ are constructed by correlating left and right image features at corresponding scales, similar to DispNetC \cite{mayer2016large}:
\begin{equation}
    \label{eq:correlation}
    \bm C^s (d, h, w) = \frac{1}{N} \langle \bm F_l^s(h, w), \bm F_r^s(h, w - d) \rangle,
\end{equation}
where $\langle\cdot,\cdot\rangle$ denotes the inner product of two feature vectors and $N$ is the channel number of extracted features. $\bm C^s (d, h, w)$ is the matching cost at location $(h, w)$ for disparity candidate $d$. The raw cost volumes $\{\bm C^s \}_{s=1}^S$ are then aggregated with several stacked Adaptive Aggregation Modules (AAModules), where an AAModule consists of $S$ adaptive Intra-Scale Aggregation (ISA) modules and an adaptive Cross-Scale Aggregation (CSA) module for $S$ pyramid levels. Finally, the predicted low-resolution disparity is hierarchically upsampled to the original resolution with the refinement modules. All disparity predictions are supervised with ground truth when training, while only the last disparity prediction is required for inference. Fig.~\ref{fig:overview} provides an overview of our proposed Adaptive Aggregation Network (AANet). In the following, we introduce the ISA and CSA modules in detail.


\subsection{Adaptive Intra-Scale Aggregation}
\label{sec:intra}

To alleviate the well-known edge-fattening issue at disparity discontinuities, we propose a sparse points based representation for efficient and flexible cost aggregation. Since the resulting formulation is similar to deformable convolution, we adopt it in our implementation. 

Specifically, for cost volume $\bm C \in \mathbb{R}^{D \times H \times W}$ at a certain scale, where $D, H, W$ represents the maximum disparity, height and width, respectively, the proposed cost aggregation strategy is defined as
\begin{equation}
    \label{eq:adaptive_sampling}
    \Tilde{\bm C}(d, \bm p) = \sum_{k=1}^{K^2} w_k \cdot \bm C (d, \bm p + \bm p_k + \Delta \bm p_k),
\end{equation}
where $\Tilde{\bm C} (d, \bm p)$ denotes the aggregated cost at pixel $\bm p$ for disparity candidate $d$, $K^2$ is the number of sampling points ($K=3$ in our paper), $w_k$ is the aggregation weight for $k$-th point, $\bm p_k$ is the fixed offset to $\bm p$ in window based cost aggregation approaches.
Our key difference from previous stereo works is that we learn additional offset $\Delta \bm p_k$ to regular sampling location $\bm p + \bm p_k$, thus enabling adaptive sampling for efficient and flexible cost aggregation, leading to high-quality results in object boundaries and thin structures.

However, in the context of learning, the spatial sharing nature of regular convolution weights $\{w_k\}_{k=1}^{K^2}$ makes themselves content-agnostic. We further learn position-specific weights $\{m_k\}_{k=1}^{K^2}$ (i.e., modulation in \cite{zhu2019deformable}, they also have effects of controlling the relative influence of the sampling points) for each pixel location $\bm p$ to achieve content-adaptive cost aggregation:
\begin{equation}
    \label{eq:content_adaptive}
    \Tilde{\bm C}(d, \bm p) = \sum_{k=1}^{K^2} w_k \cdot \bm C (d, \bm p + \bm p_k + \Delta \bm p_k) \cdot m_k.
\end{equation}
We implement Eq.~\eqref{eq:content_adaptive} with deformable convolution \cite{zhu2019deformable}, both $\Delta \bm p_k$ and $m_k$ are obtained by a separate convolution layer applied over the input cost volume $\bm C$. The original formulation of deformable convolution assumes the offsets $\Delta \bm p_k$ and weights $m_k$ are shared by each channel (i.e., disparity candidate $d$ in this paper), we further evenly divide all disparity candidates into $G$ groups, and share $\Delta \bm p_k$ and $m_k$ within each group. Dilated convolution \cite{yu2017dilated} is also used for deformable convolution to introduce more flexibility. We set $G=2$ and the dilation rate to $2$ in this paper.

We build an Intra-Scale Aggregation (ISA) module with a stack of $3$ layers and a residual connection \cite{he2016deep}. The three layers are $1\times 1, 3\times 3$ and $1\times 1$ convolutions, where the $3\times 3$ convolution is a deformable convolution. This design is similar to the bottleneck in \cite{he2016deep}, but we always keep the channels constant (equals to the number of disparity candidates). That is, we keep reasoning about disparity candidates, similar to traditional cost aggregation methods. 



\subsection{Adaptive Cross-Scale Aggregation}
\label{sec:inter}

In low-texture or textureless regions, searching the correspondence at the coarse scale can be beneficial \cite{menz2003stereoscopic}, as the texture information will be more discriminative under the same patch size when an image is downsampled. A similar observation has also been made in \cite{xu2019multi}. Therefore, multi-scale interactions
are introduced in traditional cross-scale cost aggregation algorithm \cite{zhang2014cross}.


The analysis in \cite{zhang2014cross} shows that the final cost volume is obtained through the adaptive combination of the results of cost aggregation performed at different scales (details are given in the supplementary material). We thus approximate this algorithm with
\begin{equation}
    \label{eq:cross_scale}
    \hat{\bm C}^s  = \sum_{k=1}^S f_k(\Tilde{\bm C}^k), \quad s = 1, 2, \cdots, S,
\end{equation}
where $\hat{\bm C}$ is the resulting cost volume after cross-scale cost aggregation, $\Tilde{\bm C}^k$ is the intra-scale aggregated cost volume at scale $k$, for example, with the algorithm in Sec.~\ref{sec:intra}, and $f_k$ is a general function to enable the adaptive combination of cost volumes at each scale. We adopt the definition of $f_k$ from HRNet \cite{sun2019deep}, a recent work for human pose estimation, which depends on the resolutions of cost volumes $\Tilde{\bm C}^k$ and $\hat{\bm C}^s$. Concretely, for cost volume $\hat{\bm C}^s$,
\begin{equation}
\label{eq:f_def}
    f_k = 
    \begin{cases}
    \mathcal{I}, \quad k = s, \\
    (s - k) \ \mathrm{stride-}2 \ 3 \times 3 \ \mathrm{convs}, \quad k < s, \\
    \mathrm{upsampling} \bigoplus 1 \times 1 \ \mathrm{conv}, \quad k > s,
     \end{cases}
\end{equation}
where $\mathcal{I}$ denotes the identity function, $s-k$ stride-$2$ $3 \times 3$ convolutions are used for $2^{s-k}$ times downsampling to make the resolution consistent, and $\bigoplus$ means bilinear upsampling to the same resolution first, then following a $1 \times 1$ convolution to align the number of channels. We denote this architecture as Cross-Scale Aggregation (CSA) module.

Although our CSA module is similar to HRNet\cite{sun2019deep}, they have two major differences. First, we are inspired by traditional cross-scale cost aggregation algorithm \cite{zhang2014cross} and aiming at approximating the geometric conclusion with neural network layers, while HRNet is designed for learning rich feature representations.
Moreover, the channel number (corresponding to the disparity dimension) of lower scale cost volume is \textit{halved} in our approach due to the smaller search range in coarser scales, while HRNet doubles, indicating our architecture is more efficient than HRNet.

\subsection{Adaptive Aggregation Network}
\label{sec:aanet}

The proposed ISA and CSA modules are complementary and can be integrated, resulting in our final Adaptive Aggregation Module (AAModule, see Fig.~\ref{fig:overview}). We stack six AAModules for cost aggregation, while for the first three AAModules, we simply use regular 2D convolutions for intra-scale aggregation, thus a total of nine deformable convolutions are used for cost aggregation in this paper.


Our feature extractor adopts a ResNet-like \cite{he2016deep} architecture ($40$ layers in total), in which six regular 2D convolutions are replaced with their deformable counterparts. We use Feature Pyramid Network \cite{lin2017feature} to construct feature pyramid at $1/3$, $1/6$ and $1/12$ resolutions. Two refinement modules proposed in StereoDRNet \cite{chabra2019stereodrnet} are used to hierarchically upsample the $1/3$ disparity prediction to the original resolution (i.e., upsample to $1/2$ resolution first, then to original resolution). Combining all these components leads to our final Adaptive Aggregation Network (AANet). 

\subsection{Disparity Regression}
\label{sec:estimation}

For each pixel, we adopt the \textit{soft argmin} mechanism \cite{kendall2017end} to obtain the disparity prediction $\Tilde{d}$:
\begin{equation}
    \label{eq:estimation}
    \Tilde{d} = \sum_{d=0}^{D_{\mathrm{max}} - 1} d \times \sigma (c_d),
\end{equation}
where $D_{\mathrm{max}}$ is the maximum disparity range, $\sigma$ is the softmax function, and $c_d$ is the aggregated matching cost for disparity candidate $d$. $\sigma (c_d)$ can be seen as the probability of disparity being $d$. This regression based formulation can produce sub-pixel precision and 
thus is used in this paper.

\subsection{Loss Function}
\label{sec:loss}
Our AANet is trained end-to-end with ground truth disparities as supervision. While for KITTI dataset, the high sparsity of disparity ground truth may not be very effective to drive our learning process. Inspired by the knowledge distillation in \cite{hinton2015distilling}, we propose to leverage the prediction results from a pre-trained stereo model as pseudo ground truth supervision. Specifically, we employ a pre-trained model to predict the disparity maps on the training set, and use the prediction results as pseudo labels in pixels where ground truth disparities are not available. 
We take the pre-trained GA-Net \cite{zhang2019ga} model as an example to validate the effectiveness of this strategy.


For disparity prediction $\bm D_{\mathrm{pred}}^i, i = 1, 2, \cdots, N$, it is first bilinearly upsampled to the original resolution. The corresponding loss function is defined as
\begin{align}
    L_i = & \sum_{\bm p} \bm V(\bm p) \cdot \mathcal{L} (\bm D^i_{\mathrm{pred}}(\bm p), \bm D_{\mathrm{gt}}(\bm p)) \nonumber \\
    & + (1 - \bm V(\bm p)) \cdot \mathcal{L} (\bm D_{\mathrm{pred}}^i(\bm p), \bm D_{\mathrm{pseudo}}(\bm p)),
    \label{eq:loss}
\end{align}
where $\bm V(\bm p)$ is a binary mask to denote whether the ground truth disparity for pixel $\bm p$ is available, $\mathcal{L}$ is the smooth L1 loss \cite{chang2018pyramid}, $\bm D_{\mathrm{gt}}$ is the ground truth disparity and $\bm D_{\mathrm{pseudo}}$ is the pseudo ground truth.

The final loss function is a combination of losses over all disparity predictions
\begin{equation}
    L = \sum_{i=1}^N \lambda_i \cdot L_i,
\end{equation}
where $\lambda_i$ is a scalar for balancing different terms.

\section{Experiments}

\subsection{Datasets and Evaluation Metrics}
We conduct extensive experiments on three popular stereo datasets: Scene Flow, KITTI 2012 and KITTI 2015. The Scene Flow dataset \cite{mayer2016large} is a large scale synthetic dataset
and provides dense ground truth disparity maps. 
The end-point error (EPE) and $1$-pixel error are reported on this dataset, where EPE is the mean disparity error in pixels and 1-pixel error is the average percentage of pixel whose EPE is bigger than 1 pixel. 
The KITTI 2012 \cite{geiger2012we} and KITTI 2015 \cite{menze2015object} are real-world datasets in the outdoor scenario, where only sparse ground truth is provided.
The official metrics (e.g., D1-all) in the online leader board are reported.

\subsection{Implementation Details}
We implement our approach in PyTorch \cite{paszke2019pytorch} and using Adam \cite{kingma2014adam} ($\beta_1 = 0.9, \beta_2 = 0.999$) as optimizer. 
For Scene Flow dataset, we use all training set (35454 stereo pairs) for training and evaluate on the standard test set (4370 stereo pairs). The raw images are randomly cropped to $288 \times 576$ as input. 
We train our model on 4 NVIDIA V100 GPUs for 64 epochs with a batch size of 64. The learning rate starts at 0.001 and is decreased by half every 10 epochs after 20th epoch. For KITTI dataset, we use $336 \times 960$ crop size, and first fine-tune the pre-trained Scene Flow model on mixed KITTI 2012 and 2015 training sets for 1000 epochs. The initial learning rate is 0.001 and decreased by half at 400th, 600th, 800th and 900th epochs. Then another 1000 epochs are trained on the separate KITTI 2012/2015 training set for benchmarking, with an initial learning rate of 0.0001 and same schedule as before. But only the last disparity prediction is supervised with ground truth following a similar strategy in \cite{ilg2017flownet}.
For all datasets, the input images are normalized with ImageNet mean and standard deviation statistics. We use random color augmentation and vertical flipping, and set the maximum disparity as 192 pixels. From highest scale to lowest, the loss weights in Eq.~\ref{eq:loss} are set to $\lambda_1=\lambda_2=\lambda_3=1.0, \lambda_4=2/3, \lambda_5 = 1/3$.


\begin{table}[t]
    \centering
    \begin{tabular}{lcccc}
    \toprule
    \multirow{2}{*}[-2pt]{Method} & \multicolumn{2}{c}{Scene Flow} & \multicolumn{2}{c}{KITTI 2015} \\
    \cmidrule(lr){2-3} \cmidrule(lr){4-5} \\
    \addlinespace[-10pt]
    & EPE & $> 1$px & EPE  & D1-all \\
    \midrule
    w/o ISA \& CSA & 1.10 & 10.9 & 0.75 & 2.63 \\
    w/o ISA & 0.97 & 10.1 & 0.70 & {\bf 2.22} \\
    w/o CSA & 0.99 & 10.1 & 0.69 & 2.31 \\
    AANet & {\bf 0.87} & {\bf 9.3} & {\bf 0.68} & {2.29} \\
    \bottomrule
    \end{tabular}
    \caption{Ablation study of ISA and CSA modules. The best performance is obtained by integrating these two modules.}
    \label{tab:ablation}
\end{table}

\begin{figure}[t]
\centering
\setlength{\tabcolsep}{0.5pt}

{\renewcommand{\arraystretch}{0.3} 

\begin{tabular}{lcccc}

\rotatebox{90}{{\tiny  \qquad \quad Image}} &
\includegraphics[width=0.33\linewidth]{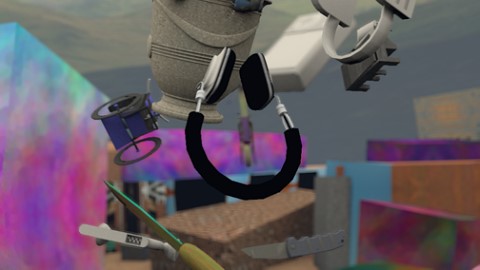} &
\includegraphics[width=0.33\linewidth]{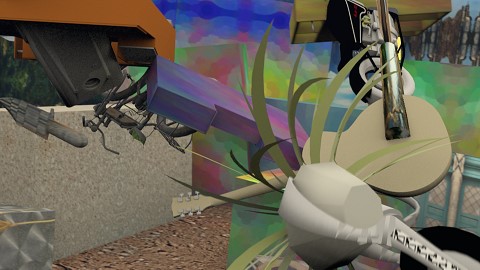} &
\includegraphics[width=0.33\linewidth]{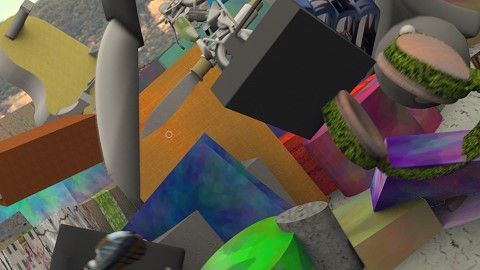} \\

\rotatebox{90}{{\tiny  \quad w/o ISA \& CSA }} &
\includegraphics[width=0.33\linewidth]{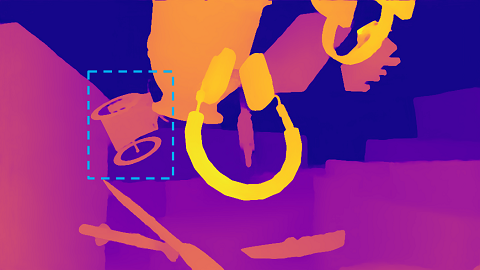} &
\includegraphics[width=0.33\linewidth]{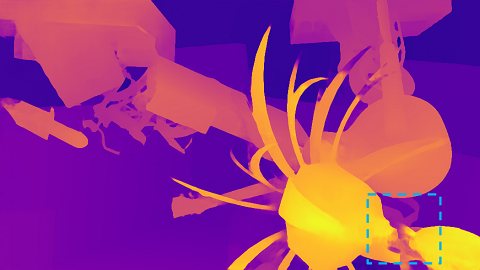} &
\includegraphics[width=0.33\linewidth]{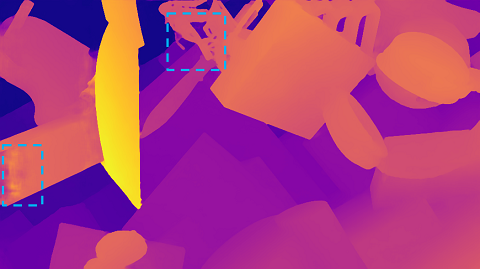} \\

\rotatebox{90}{{\tiny  \qquad \quad AANet}} &
\includegraphics[width=0.33\linewidth]{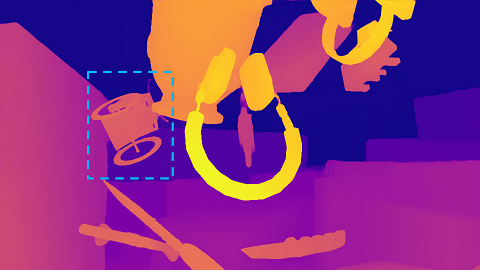} &
\includegraphics[width=0.33\linewidth]{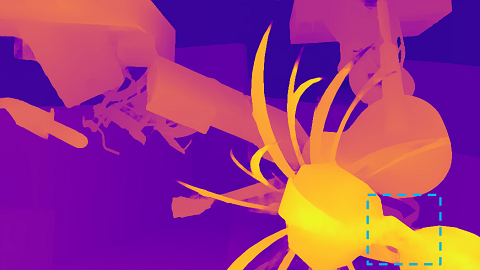} &
\includegraphics[width=0.33\linewidth]{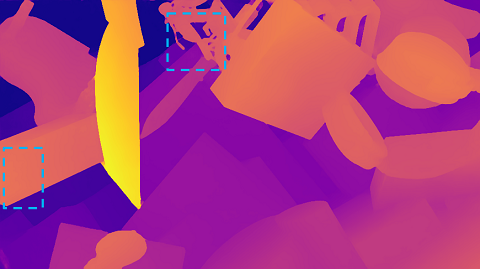} \\

\rotatebox{90}{{\tiny  \qquad \qquad GT}} &
\includegraphics[width=0.33\linewidth]{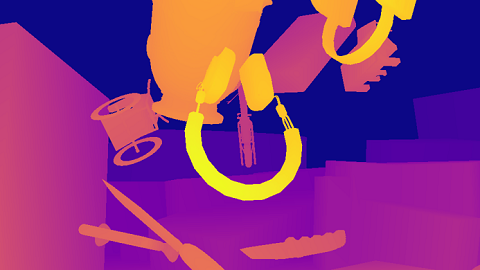} &
\includegraphics[width=0.33\linewidth]{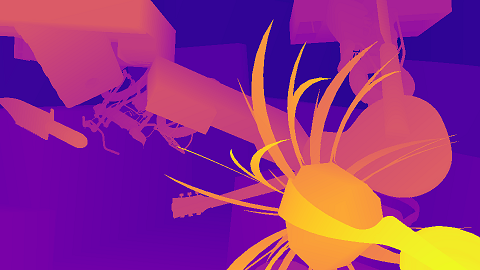} &
\includegraphics[width=0.33\linewidth]{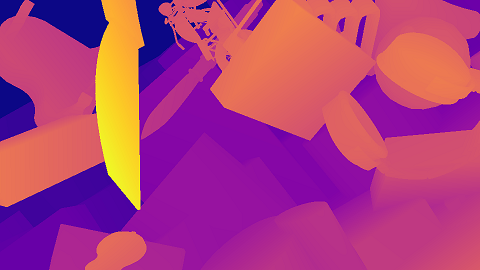} \\

\end{tabular}
}
\caption{Visual comparisons of ablation study on Scene Flow test set. Our AANet produces sharper results in thin structures and better predictions in textureless regions.}
\label{fig:ablation_vis}
\end{figure}

\begin{table*}[!htbp]
\centering

\begin{tabular}{lcccccrrrr}
\toprule
Method & \#3D Convs & \#DConvs & \#CSA & EPE & $> 1$px & Params & FLOPs & Memory & Time (ms) \\
\midrule
StereoNet \cite{khamis2018stereonet} & 4 & 0 & 0 & 1.10 & - & 0.62M & 106.89G & 1.41G & 23 \\
StereoNet-AA & 0 & 4 & 0 & {\bf 1.08 } & 12.9 & {\bf 0.53M} & {\bf 88.17G} & {\bf 1.38G} & {\bf 17} \\
\midrule
GC-Net \cite{kendall2017end} & 19 & 0 & 0 & 2.51 & 16.9 & 2.85M & 1754.10G & 21.52G & 3731 \\
GC-Net-AA & 0 & 9 & 6 & {\bf 0.98} & {\bf 10.8} & {\bf 2.15M} & {\bf 212.59G} & {\bf 1.97G} & {\bf 91}  \\
\midrule
PSMNet \cite{chang2018pyramid} & 25 & 0 & 0 & 1.09 & 12.1 & 5.22M & 613.90G & 4.08G & 317 \\
PSMNet-AA & 0 & 9 & 6 & {\bf 0.97} & {\bf 10.2} & {\bf 4.15M} & {\bf 208.73G} & {\bf 1.58G} & {\bf 77} \\
\midrule
GA-Net \cite{zhang2019ga} & 15 & 0 & 0 & {\bf 0.84} & 9.9 & 4.60M & 1439.57G & 6.23G & 2211 \\
GA-Net-AA & 0 & 14 & 6 & 0.87 & {\bf 9.2} & {\bf 3.68M} & {\bf 119.64G} & {\bf 1.63G} & {\bf 57} \\
\bottomrule
\end{tabular}
\caption{Comparisons with four representative stereo models: StereoNet, GC-Net, PSMNet and GA-Net. We replace the 3D convolutions in cost aggregation stage with our proposed architectures and denote the resulting model with suffix AA. Our method not only obtains clear performance improvements (except GA-Net has lower EPE), but also shows fewer parameters, less computational cost and memory consumption, while being significantly faster than top-performing models ($41\times$ than GC-Net, $4\times$ than PSMNet and $38\times$ than GA-Net). The comparison with StereoNet indicates that our method can also be a valuable way to improve the performance of existing fast stereo models. ``DConvs'' is short for deformable convolutions.}
\label{tab:compare_3d}
\end{table*}

\subsection{Analysis}

To validate the effectiveness of each component proposed in this paper, we conduct controlled experiments on Scene Flow test set and KITTI 2015 validation set (the KITTI 2015 training set is split into 160 pairs for training and 40 pairs for validation).


{\bf Ablation Study.} As shown in Tab.~\ref{tab:ablation}, removing the proposed ISA or CSA module leads to clear performance drop. The best performance is obtained by integrating these two modules, which are designed to be complementary in principle. Fig.~\ref{fig:ablation_vis} further shows the visual comparison results. Our full model produces better disparity predictions in thin structures and textureless regions, demonstrating the effectiveness of the proposed method.

\begin{figure}[t]
    \centering
    \begin{subfigure}{0.48\linewidth}
        \includegraphics[width=\linewidth]{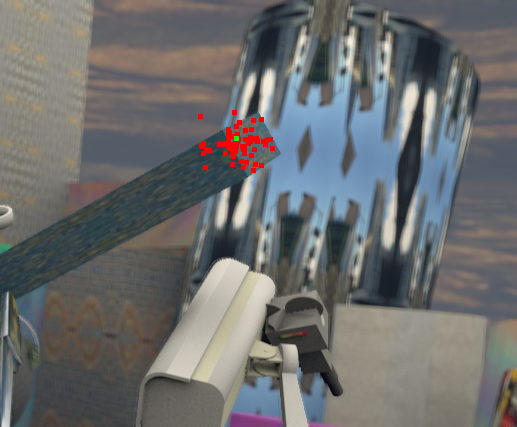}
        \caption{object boundary}
        \label{fig:vis_boundary}
    \end{subfigure}
    \begin{subfigure}{0.48\linewidth}
        \includegraphics[width=\linewidth]{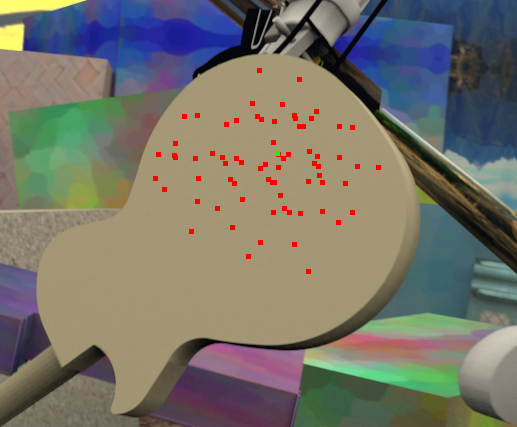}
        \caption{textureless region}
        \label{fig:vis_textureless}
    \end{subfigure}
    \caption{Visualization of sampling points (red points) in two challenging regions (green points).
    In object boundary (a), the sampling points tend to focus on similar disparity regions. While for large textureless region (b), they are more discretely distributed to sample from a large context.
    }
    \label{fig:vis_sampling_points}
\end{figure}

\begin{figure}[t]
\centering
\setlength{\tabcolsep}{0.5pt}

{\renewcommand{\arraystretch}{0.3} 

\begin{tabular}{lccc}

\rotatebox{90}{{\tiny \qquad Image}} &
\includegraphics[width=0.48\linewidth]{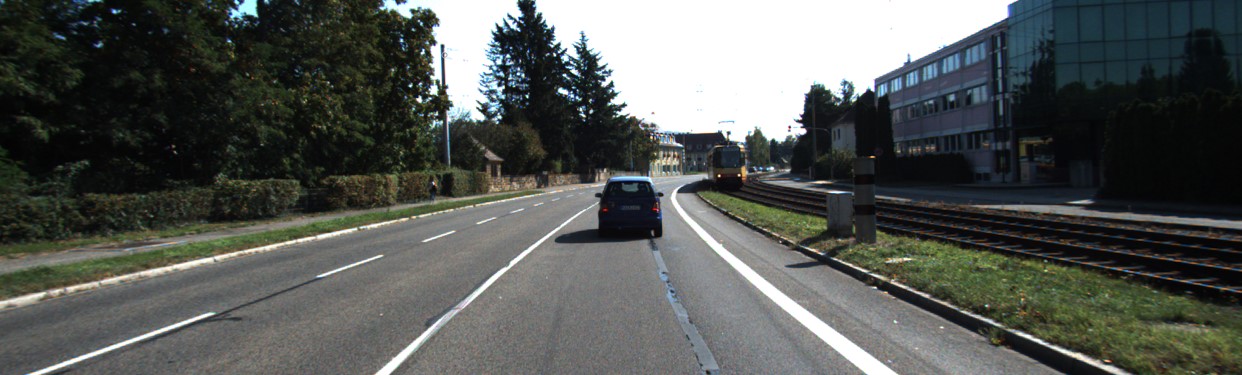} &
\includegraphics[width=0.48\linewidth]{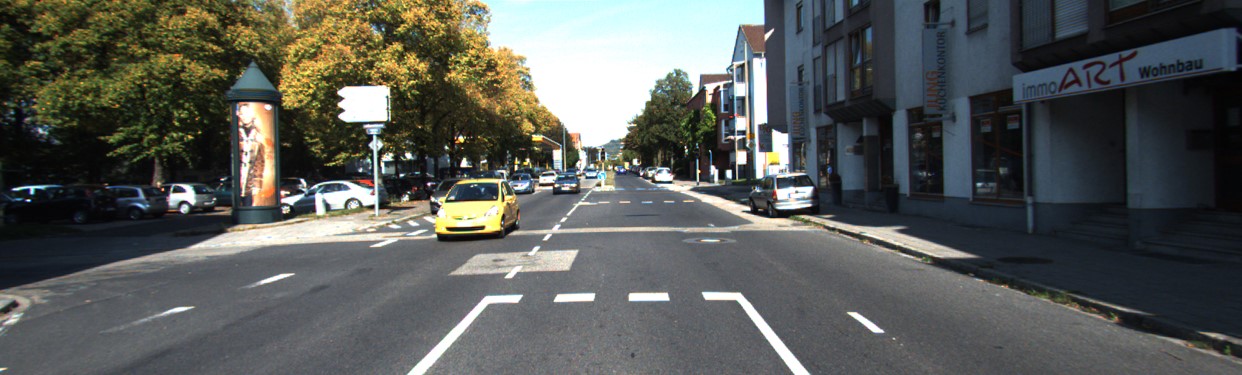} \\

\rotatebox{90}{{\tiny  \quad w/o pseudo gt}} &
\includegraphics[width=0.48\linewidth]{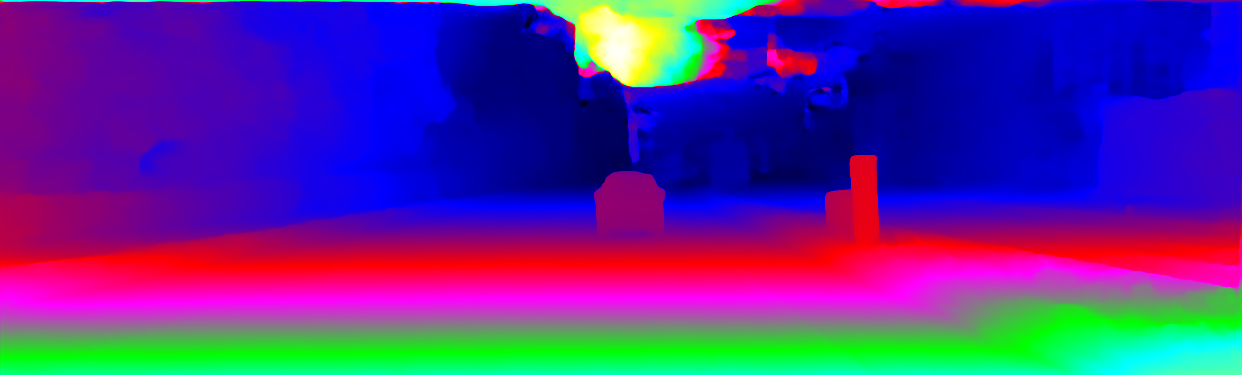} &
\includegraphics[width=0.48\linewidth]{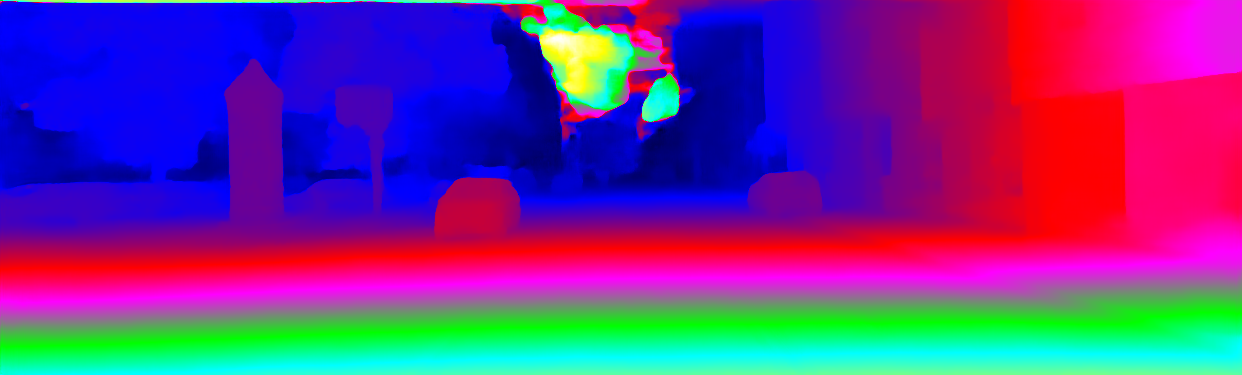} \\

\rotatebox{90}{{\tiny  \quad w/ pseudo gt}} &
\includegraphics[width=0.48\linewidth]{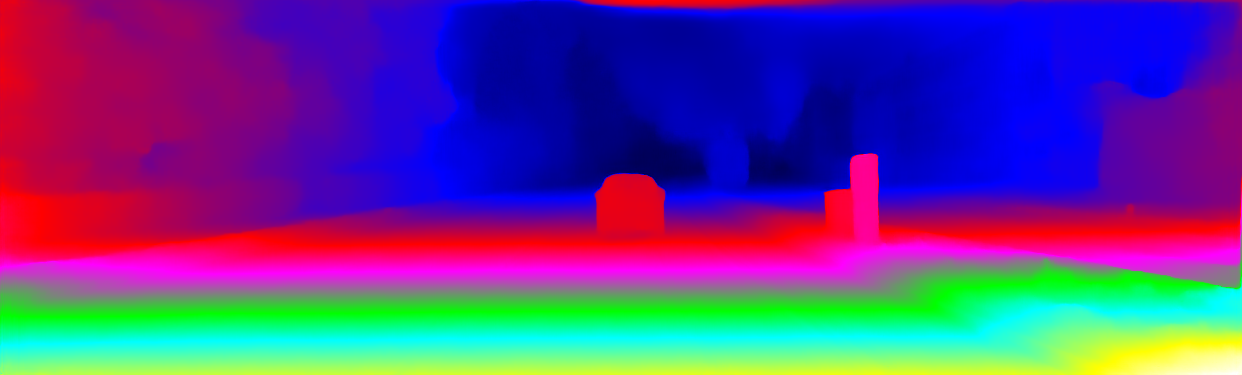} &
\includegraphics[width=0.48\linewidth]{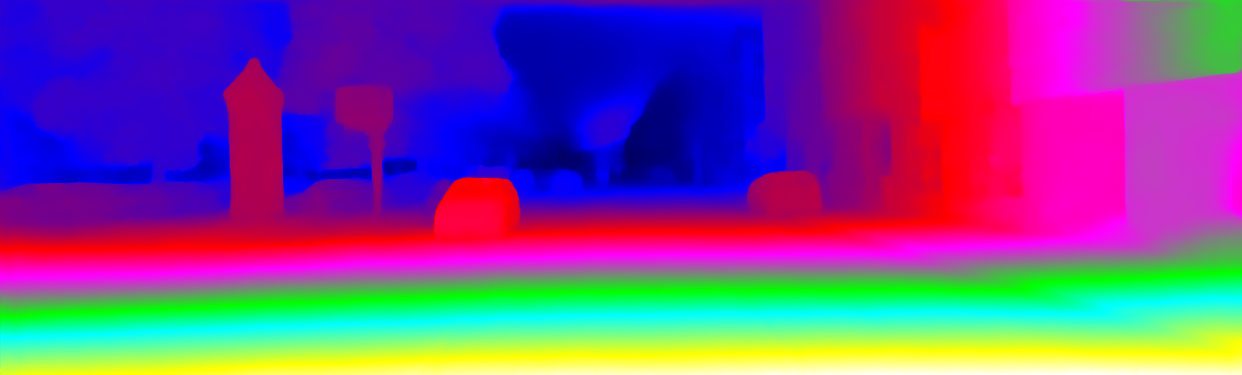} \\


\end{tabular}
}
\caption{Visualization of disparity prediction results on KITTI 2015 validation set. Leveraging pseudo ground truth as additional supervision helps reduce the artifacts in regions where ground truth disparities are not available, e.g., the sky region.}
\label{fig:pseudo_gt}
\end{figure}

{\bf Sampling Points Visualization.} To better understand our proposed adaptive intra-scale cost aggregation algorithm, we visualize the sampling locations in two challenging regions. As illustrated in Fig.~\ref{fig:vis_sampling_points}, for pixel in object boundary (Fig.~\ref{fig:vis_boundary}), the sampling points tend to focus on similar disparity regions. While for large textureless region (Fig.~\ref{fig:vis_textureless}), a large context is usually required to obtain reliable matching due to lots of local ambiguities. Our method can successfully adapt the sampling locations to these regions, validating that the proposed adaptive aggregation method can not only dynamically adjust the sampling locations, but also enables sampling from a large context.

{\bf Pseudo Ground Truth Supervision.} Fig.~\ref{fig:pseudo_gt} shows the visual results on KITTI 2015 validation set. We empirically find that leveraging the prediction results from a pre-trained GA-Net \cite{zhang2019ga} model helps reduce the artifacts in regions where ground truth disparities are not available, e.g., the sky region. Quantitatively, the D1-all error metric decreases from 2.29 to 2.15, while the EPE increases from 0.68 to 0.69. The possible reason might be that the validation set is too small to make the results unstable. Similar phenomenon has also been noticed in \cite{guo2019group}. However, the qualitative results indicate that our proposed strategy can be an effective way to handle highly sparse ground truth data.

\begin{figure}[t]
\centering
\setlength{\tabcolsep}{0.5pt}

{
\renewcommand{\arraystretch}{0.3} 

\begin{tabular}{ccc}

Image & PSMNet & AANet \\
\includegraphics[width=0.33\linewidth]{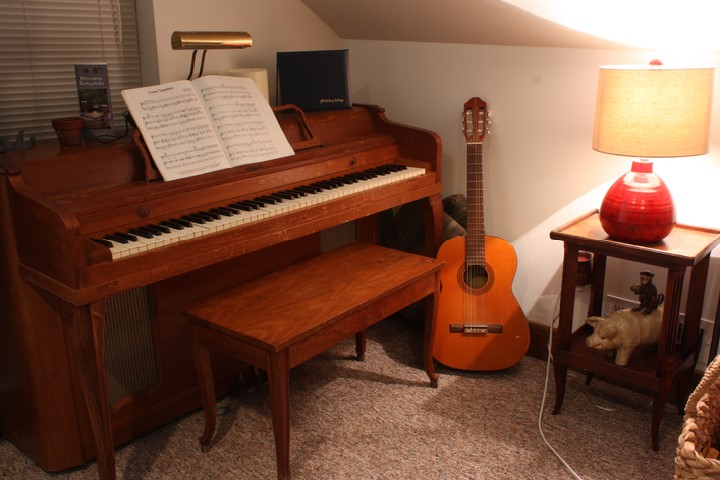} &
\includegraphics[width=0.33\linewidth]{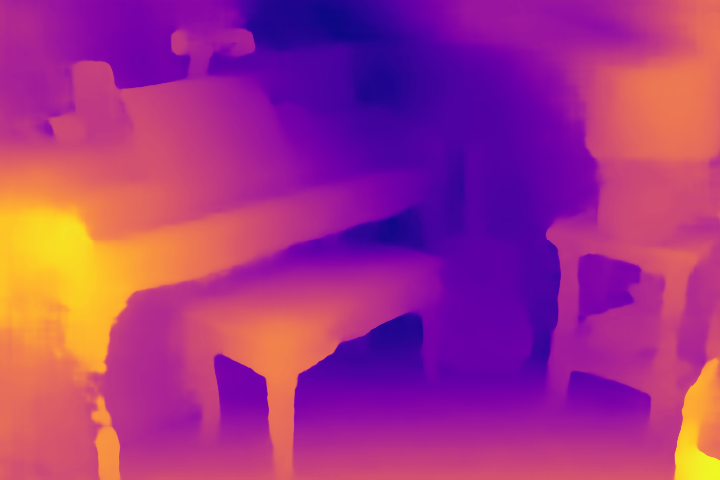} &
\includegraphics[width=0.33\linewidth]{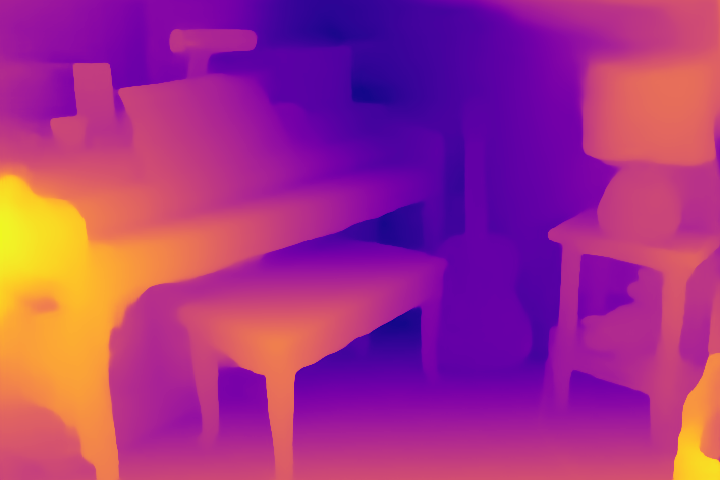} \\
\includegraphics[width=0.33\linewidth]{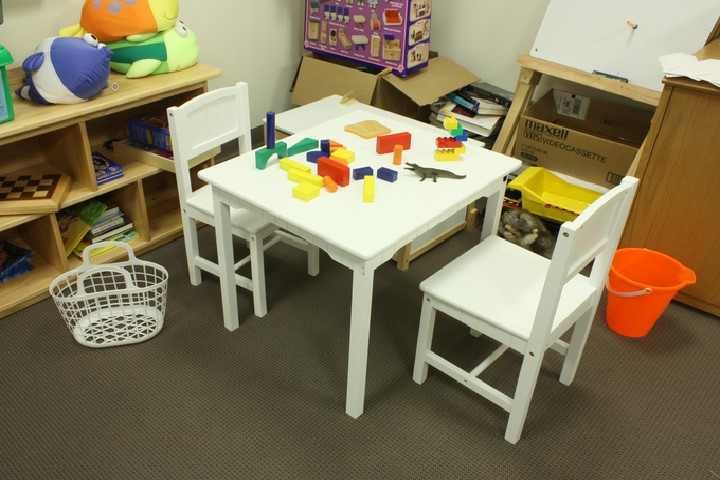} &
\includegraphics[width=0.33\linewidth]{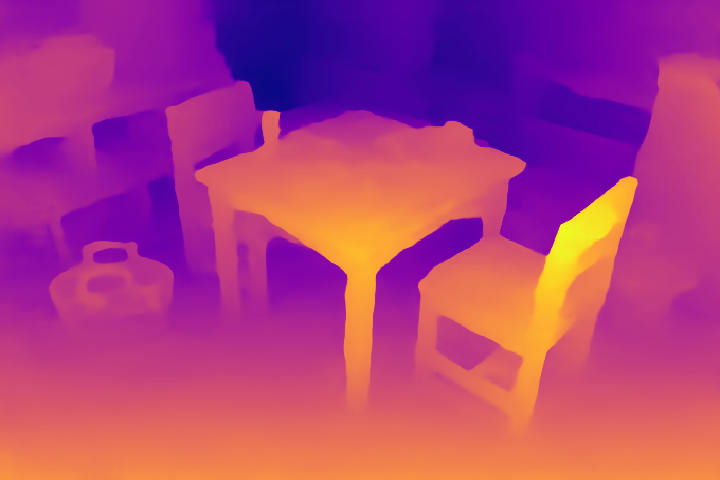} &
\includegraphics[width=0.33\linewidth]{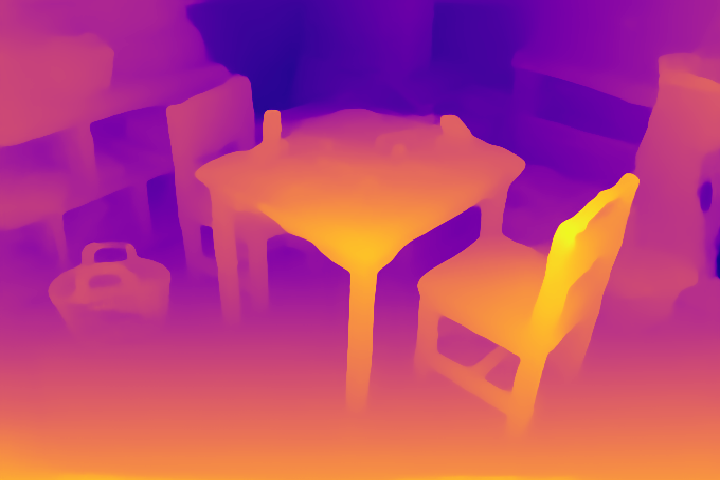} \\

\end{tabular}
}
\caption{Generalization on Middlebury 2014 dataset. Our AANet produces sharper object boundaries and better preserves the overall structures than PSMNet.}
\label{fig:middlebury}
\end{figure}

{\bf Generalization.} We further test the generalization ability of our method on Middlebury 2014 dataset \cite{scharstein2014high}. Specifically, we directly use our KITTI fine-tuned model to predict the disparity map, no additional training is done on Middlebury. Fig.~\ref{fig:middlebury} shows the results. Compared with the popular PSMNet \cite{chang2018pyramid} model, our AANet produces sharper object boundaries and better preserves the overall structures.

\begin{table*}[!htbp]
    \centering
    \footnotesize
    
    \begin{tabular}{ccccccccc}
    \toprule
    Method & GC-Net \cite{kendall2017end} & PSMNet \cite{chang2018pyramid} & GA-Net \cite{zhang2019ga} & DeepPruner-Best \cite{duggal2019deeppruner} & DispNetC \cite{mayer2016large} & StereoNet \cite{khamis2018stereonet} &
    AANet & AANet+ \\
    \midrule
    EPE & 2.51 & 1.09 & 0.84 & 0.86 & 1.68 & 1.10 & 0.87 & {\bf 0.72} \\
    Time (s) & 0.9 & 0.41 & 1.5 & 0.182 & 0.06 & {\bf 0.015} & 0.068 & 0.064 \\
    \bottomrule
     
    \end{tabular}
    \caption{Evaluation results on Scene Flow test set. Our method not only achieves state-of-the-art performance but also runs significantly faster than existing top-performing methods.}
    \label{tab:sceneflow}
\end{table*}

\begin{table}[!htbp]
\centering
\setlength{\tabcolsep}{5pt}
\footnotesize

\begin{tabular}{lccccc}
\toprule
\multirow{2}{*}{Method} &
\multicolumn{2}{c}{KITTI 2012} &
\multicolumn{2}{c}{KITTI 2015} &
\multirow{2}{*}{\thead{Time \\(s)}} \\
\cmidrule(lr){2-3} \cmidrule(lr){4-5} \\
\addlinespace[-10pt]
& Out-Noc & Out-All & D1-bg & D1-all & \\
\midrule
MC-CNN \cite{zbontar2015computing} & 2.43 & 3.63 & 2.89 & 3.89 & 67 \\
GC-Net \cite{kendall2017end} & 1.77 & 2.30 & 2.21 & 2.87 & 0.9 \\
PSMNet \cite{chang2018pyramid} & 1.49 & 1.89 & 1.86 & 2.32 & 0.41 \\
DeepPruner-Best \cite{duggal2019deeppruner} & - & - & 1.87 & 2.15 & 0.182 \\
iResNet-i2 \cite{liang2018learning} & 1.71 & 2.16 & 2.25 & 2.44 & 0.12 \\
HD$^3$ \cite{yin2019hierarchical} & 1.40 & 1.80 & 1.70 & 2.02 & 0.14 \\
GwcNet \cite{guo2019group} & {\bf 1.32} & {\bf 1.70} & 1.74 & 2.11 & 0.32 \\
GA-Net \cite{zhang2019ga} & 1.36 & 1.80 & {\bf 1.55} & {\bf 1.93} & 1.5 \\
AANet+ & 1.55 & 2.04 & 1.65 & 2.03  & {\bf 0.06} \\
\midrule
StereoNet \cite{khamis2018stereonet} & 4.91 & 6.02 & 4.30 & 4.83 & {\bf 0.015} \\
MADNet \cite{tonioni2019real} & - & - & 3.75 & 4.66 & 0.02 \\
DispNetC \cite{mayer2016large} & 4.11 & 4.65 & 4.32 & 4.34 & 0.06 \\
DeepPruner-Fast \cite{duggal2019deeppruner} & - & - & 2.32 & 2.59 & 0.061 \\
AANet & {\bf 1.91} & {\bf 2.42} & {\bf 1.99} & {\bf 2.55} & 0.062 \\
\bottomrule
\end{tabular}
\caption{Benchmark results on KITTI 2012 and KITTI 2015 test sets. Our AANet+ model achieves competitive results among existing top-performing methods while being considerably faster. Compared with other fast models, our AANet is much more accurate.}
\label{tab:kitti}
\end{table}

\subsection{Comparison with 3D Convolutions}

To demonstrate the superiority of our proposed cost aggregation method over commonly used 3D convolutions, we conduct extensive experiments on the large scale Scene Flow dataset. 

{\bf Settings.} We mainly compare with four representative stereo models: the first 3D convolution based model GC-Net \cite{kendall2017end}, real-time model StereoNet \cite{khamis2018stereonet}, previous and current state-of-the-art models PSMNet \cite{chang2018pyramid} and GA-Net \cite{zhang2019ga}. For fair comparisons, we use similar feature extractors with them. Specifically, StereoNet uses $8\times$ downsampling for fast speed while we use $4\times$; five regular 2D convolutions in GA-Net are replaced with their deformable counterparts; for GC-Net and PSMNet, the feature extractors are exactly the same. 
We replace the 3D convolutions in cost aggregation stage with our proposed AAModules, and denote the resulting model with suffix AA. We integrate all these models in a same framework and measure the inference time with $576 \times 960$ resolution on a single NVIDIA V100 GPU.

{\bf Results.} Tab.~\ref{tab:compare_3d} shows the comprehensive comparison metrics/statistics. 
To achieve fast speed, StereoNet \cite{khamis2018stereonet} uses $8\times$ downsampling to build a very low-resolution cost volume, while at the cost of sacrificing accuracy. But thanks to our efficient adaptive aggregation architecture, we are able to directly aggregate the $1/4$ cost volume with even less computation while being more accurate and faster, indicating that our method can be a valuable way to improve the performance of existing fast stereo models. Compared with top-performing stereo models GC-Net \cite{kendall2017end}, PSMNet \cite{chang2018pyramid} and GA-Net \cite{zhang2019ga}, we not only obtain clear performance improvements (except GA-Net has lower EPE than ours), but also show fewer parameters, less computational cost and memory consumption, while being significantly faster ($41\times$ than GC-Net, $4\times$ than PSMNet and $38\times$ than GA-Net), demonstrating the high efficiency of our method compared with commonly used 3D convolutions.

{\bf Complexity Analysis.} 2D stereo methods use simple feature correlation to build a 3D cost volume ($D \times H \times W$) while 3D methods use concatenation thus a 4D cost volume is built ($C \times D \times H \times W$), where $C, D, H, W$ denotes channels after feature concatenation, maximum disparity, height and width, respectively. $C$ usually equals to $64$ for 3D convolutions based methods and $D = 64$ for $1/3$ resolution cost volume. Supposing the output cost volume has the same size as input and the kernel size of a convolution layer is $K$ ($K=3$ usually), then the computational complexity of a 3D convolution layer is $\mathcal{O} (K^3C^2DHW)$. In contrast, the complexity of a deformable convolution layer is $\mathcal{O} (K^2D^2HW + 3K^4DHW + 3K^2DHW)$.
Therefore, the computational complexity of a deformable convolution layer is less than $1/130$ of a 3D convolution layer.

\subsection{Benchmark Results}

\begin{figure}[t]
\centering
\setlength{\tabcolsep}{0.5pt}

{\renewcommand{\arraystretch}{0.3} 

\begin{tabular}{lcc}

\rotatebox{90}{{\tiny \quad  Image}} &
\includegraphics[width=0.48\linewidth]{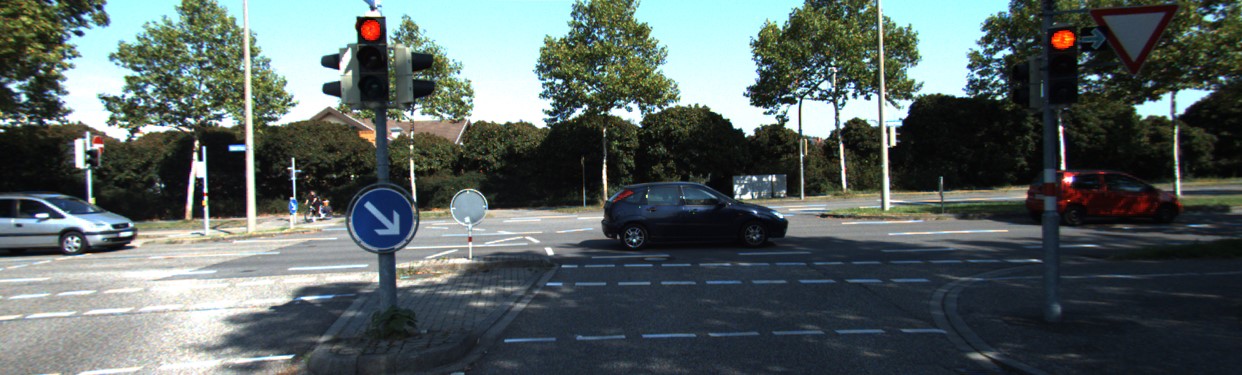} &
\includegraphics[width=0.48\linewidth]{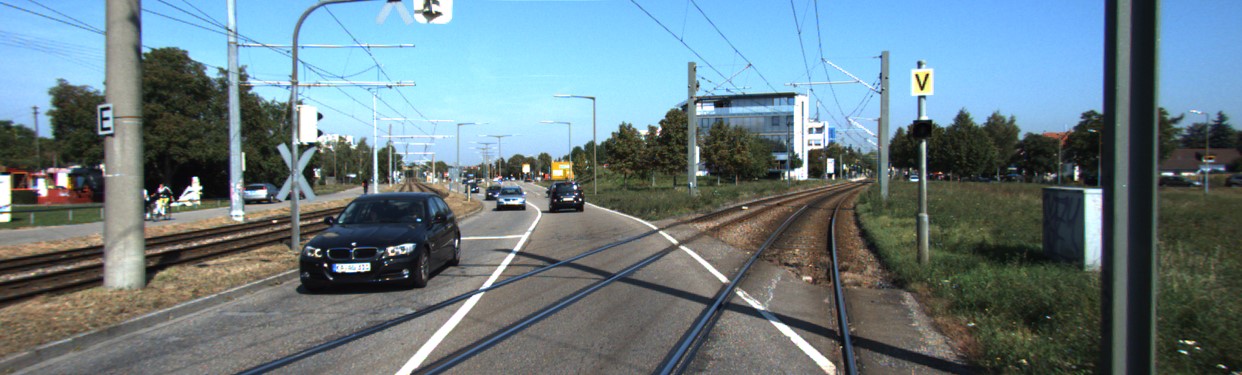} \\

\rotatebox{90}{{\tiny \quad  DispNetC }} &
\includegraphics[width=0.48\linewidth]{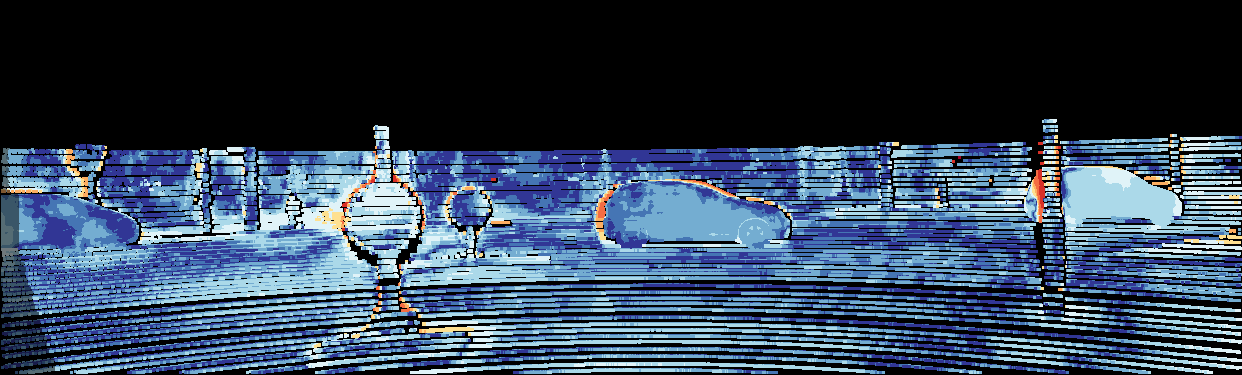} &
\includegraphics[width=0.48\linewidth]{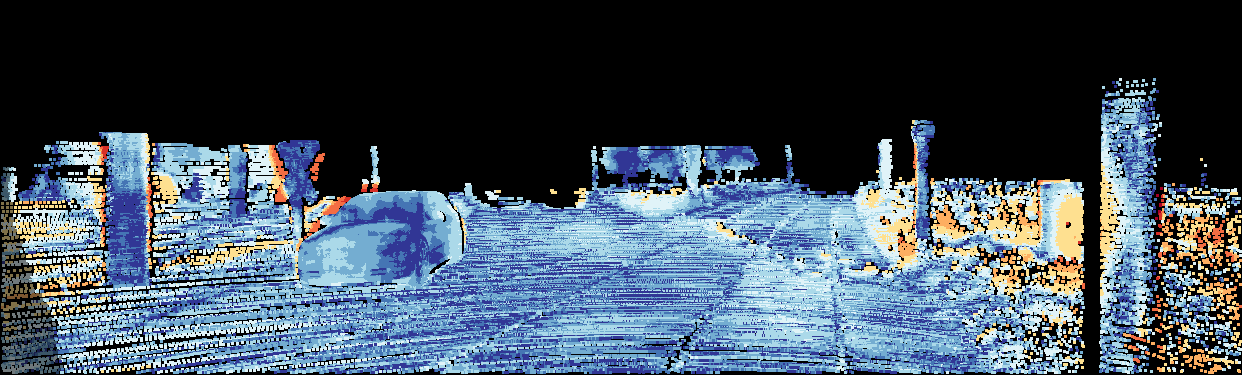} \\

\rotatebox{90}{{\tiny \quad  PSMNet }} &
\includegraphics[width=0.48\linewidth]{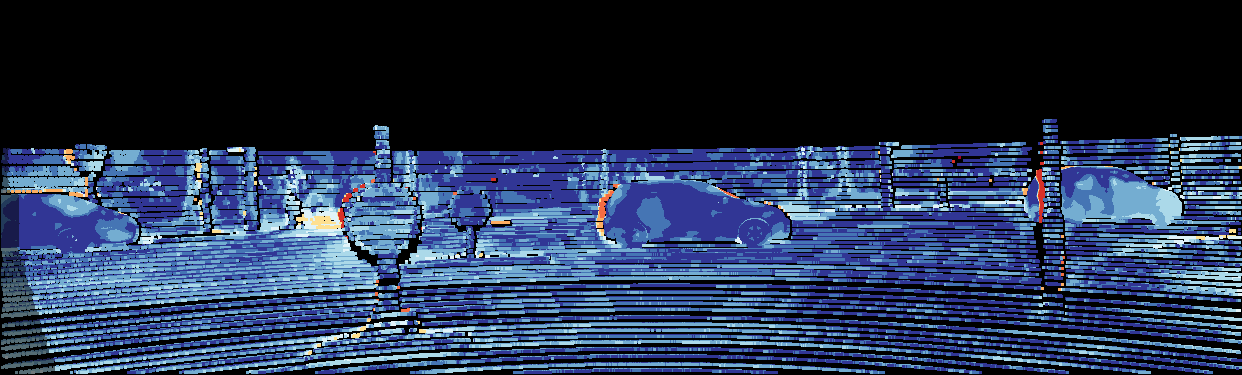} &
\includegraphics[width=0.48\linewidth]{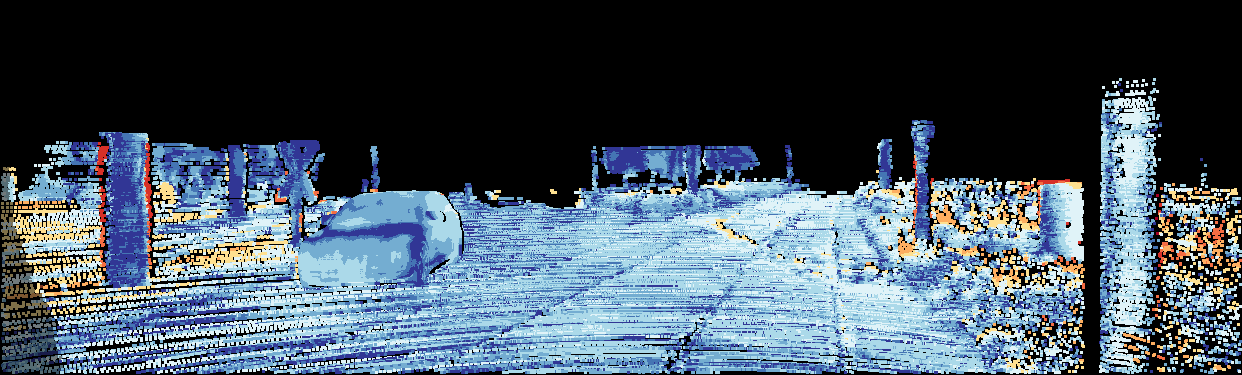} \\

\rotatebox{90}{{\tiny \quad AANet}} &
\includegraphics[width=0.48\linewidth]{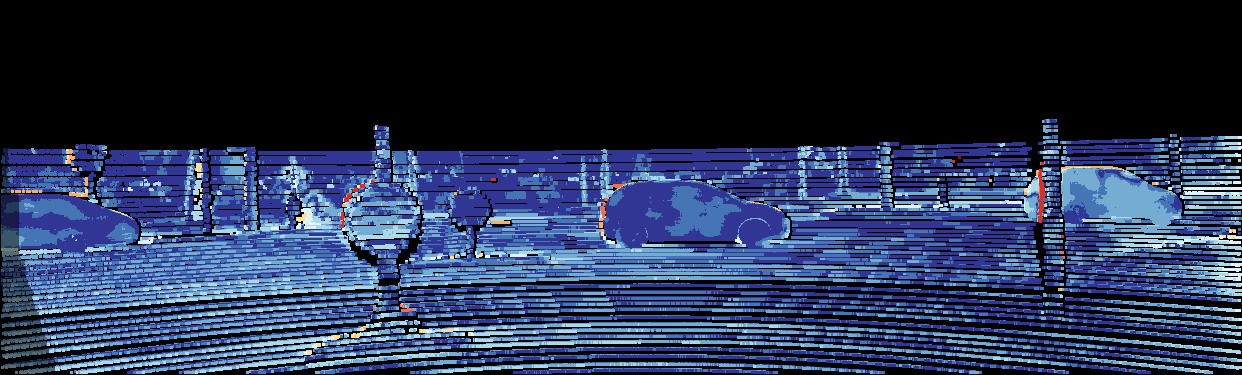} &
\includegraphics[width=0.48\linewidth]{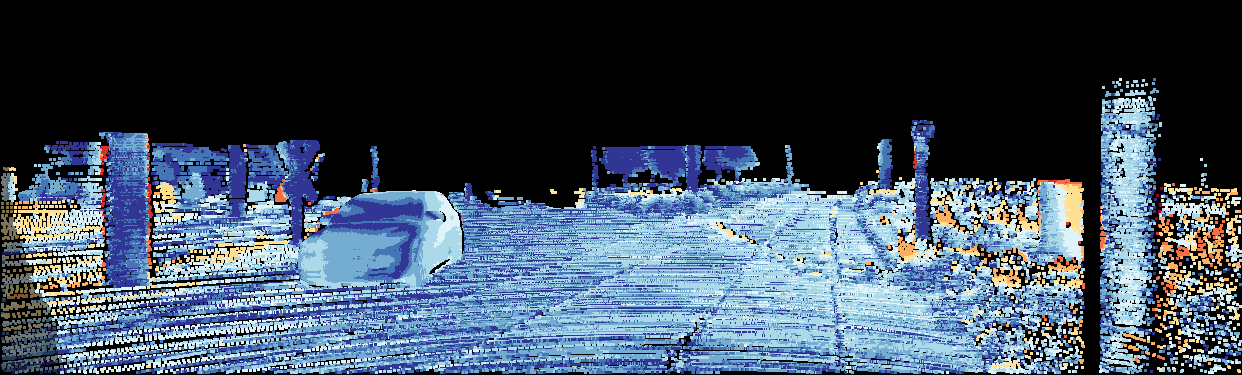} \\

\end{tabular}
}
\caption{Visualization of disparity prediction error on KITTI 2015 test set (red and yellow denote large errors). Our method produces better results in object boundaries. Best viewed enlarged. }
\label{fig:compare_kitti15}
\end{figure}

For benchmarking, we build another model variant AANet+. Specifically, the AANet+ model is built by replacing the refinement modules in the GA-Net-AA (see Tab.~\ref{tab:compare_3d}) model with hourglass networks, and five regular 2D convolutions in each refinement module are replaced with their deformable counterparts. We note that our AANet+ has more parameters than AANet (8.4M vs.\ 3.9M), but it still enjoys fast speed. Tab.~\ref{tab:sceneflow} shows the evaluation results on Scene Flow test set. Our method not only achieves state-of-the-art results, but also runs significantly faster than existing top-performing methods. The evaluation results on KITTI 2012 and KITTI 2015 benchmarks are shown in Tab.~\ref{tab:kitti}. Compared with other fast models, our AANet is much more accurate. The AANet+ model achieves competitive results among existing top-performing methods while being considerably faster. We also note that HD$^3$\cite{yin2019hierarchical} has more than $4\times$ parameters than our AANet+ (39.1M vs.\ 8.4M), and our AANet+ performs much better than previous robust vision challenge winner\footnote{\url{http://www.robustvision.net/rvc2018.php}}, iResNet-i2\cite{liang2018learning}, demonstrating that our method achieves a better balance between accuracy and speed. Fig.~\ref{fig:compare_kitti15} further visualizes the disparity prediction error on KITTI 2015 test set. Our AANet produces better results in object boundaries, validating the effectiveness of our proposed adaptive aggregation algorithm.

\section{Conclusion}


We have presented an efficient architecture for cost aggregation, and demonstrated its superiority over commonly used 3D convolutions by high efficiency and competitive performance on both Scene Flow and KITTI datasets. Extensive experiments also validate the generic applicability of the proposed method. An interesting future direction would be extending our method to other cost volume based tasks, e.g., high-resolution stereo matching \cite{yang2019hierarchical}, multi-view stereo \cite{yao2018mvsnet} and optical flow estimation \cite{sun2018pwc}. We also hope our lightweight design can be beneficial for downstream tasks, e.g., stereo based 3D object detection \cite{wang2019pseudo}.

{\bf Acknowledgements.} We thank anonymous reviewers for their constructive comments. This work was supported by the National Natural Science Foundation of China (No. 61672481) and Youth Innovation Promotion Association CAS (No. 2018495).



\section*{Appendix}

We briefly review traditional cross-scale cost aggregation algorithm \cite{zhang2014cross} to make this paper self-contained.

For cost volume $\bm C \in \mathbb{R}^{D \times H \times W}$, \cite{zhang2014cross} reformulates the local cost aggregation from an optimization perspective:
\begin{equation}
    \Tilde{\bm C} (d, \bm p) = \argmin_z \sum_{\bm q \in N (\bm p)} w(\bm p, \bm q) \Vert z - \bm C(d, \bm q) \Vert^2,
    \label{eq:single_scale}
\end{equation}
where  $\Tilde{\bm C} (d, \bm p)$ denotes the aggregated cost at pixel $\bm p$ for disparity candidate $d$, pixel $\bm q$ belongs to the neighbors $N(\bm p)$ of $\bm p$, and $w$ is the weighting function to measure the similarity of pixel $\bm p$ and $\bm q$. The solution of this weighted least square problem \eqref{eq:single_scale} is
\begin{equation}
    \Tilde{\bm C} (d, \bm p) = \sum_{\bm q \in N(\bm p)} w (\bm p, \bm q) \bm C(d, \bm q).
\end{equation}
Thus, different local cost aggregation methods can be reformulated within a unified framework. 

Without considering multi-scale interactions, the multi-scale version of Eq.~\eqref{eq:single_scale} can be expressed as
\begin{equation}
    \Tilde{\bm v} = \argmin_{\{z^s\}_{s=1}^S} \sum_{s = 1}^S \sum_{\bm q^s \in N(\bm p^s)} w (\bm p^s, \bm q^s) \Vert z^s - \bm C^s(d^s, \bm q^s) \Vert^2, 
    \label{eq:multi-scale}
\end{equation}
where $\bm p^s$ and $d^s$ denote pixel and disparity at scale $s$, respectively, and $\bm p^{s+1} = \bm p^s / 2$, $d^{s+1} = d^s / 2$, $\bm p^1 = \bm p$ and $d^1 = d$.
The aggregated cost at each scale is denoted as
\begin{equation}
    \Tilde{\bm v} = [\Tilde{\bm C}^1 (d^1, \bm p^1), \Tilde{\bm C}^2 (d^2, \bm p^2), \cdots, \Tilde{\bm C}^S (d^S, \bm p^S)]^T.
    \label{eq:tilde_v}
\end{equation}
The solution of Eq.~\eqref{eq:multi-scale} is obtained by performing cost aggregation at each scale independently:
\begin{align}
    \Tilde{\bm C}^s (d^s, \bm p^s) = & \sum_{\bm q^s \in N(\bm p^s)} w(\bm p^s, \bm q^s) \bm C^s (d^s, \bm q^s), \nonumber \\
    & \quad s = 1, 2, \cdots, S.
\end{align}

By enforcing the inter-scale consistency on the cost volume, we can obtain the following optimization problem:
\begin{align}
    \hat{\bm v} = & \argmin_{\{z^s\}_{s=1}^S} \left( \sum_{s = 1}^S \sum_{\bm q^s \in N(\bm p^s)} w (\bm p^s, \bm q^s) \Vert z^s - \bm C^s(d^s, \bm q^s) \Vert^2 \right. \nonumber \\
    & \left. + \lambda \sum_{s=2}^S \Vert z^s - z^{s-1} \Vert^2 \right), 
    \label{eq:final}
\end{align}
where $\lambda$ is a parameter to control the regularization strength, and $\hat{\bm v}$ is denoted as
\begin{equation}
    \hat{\bm v} = [\hat{\bm C}^1 (d^1, \bm p^1), \hat{\bm C}^2 (d^2, \bm p^2), \cdots, \hat{\bm C}^S (d^S, \bm p^S)]^T.
    \label{eq:hat_v}
\end{equation}
The optimization problem \eqref{eq:final} is convex and can be solved analytically (see details in \cite{zhang2014cross}). The solution can be expressed as 
\begin{equation}
    \hat{\bm v} = \bm P \Tilde{\bm v},
    \label{eq:conclusion}
\end{equation}
where $\bm P$ is an $S \times S$ matrix. That is, the final cost volume is obtained through the adaptive combination of the results of cost aggregation performed at different scales. 

Inspired by this conclusion, we design our cross-scale cost aggregation architecture as
\begin{equation}
    \hat{\bm C}^s = \sum_{k=1}^S f_k (\Tilde{\bm C}^k), \quad s = 1, 2, \cdots, S,
\end{equation}
where $f_k$ is defined by neural network layers.

{\small
\bibliographystyle{ieee_fullname}
\bibliography{main}
}

\end{document}